# Collaborative Design of Artificial Intelligence-Enhanced Learning Activities

Margarida Romero
Université Côte d'Azur, France

https://orcid.org/0000-0003-3356-8121

## Abstract

Artificial intelligence has accelerated innovations in different aspects of citizens' lives. Many contexts have already addressed technology-enhanced learning, but educators at different educational levels now need to develop AI literacy and the ability to integrate appropriate AI usage into their teaching. We take into account this objective, along with the creative learning design, to create a formative intervention that enables preservice teachers, in-service teachers, and EdTech specialists to effectively incorporate AI into their teaching practices. We developed the formative intervention with Terra Numerica and Maison de l'Intelligence Artificielle in two phases in order to enhance their understanding of AI and foster its creative application in learning design. Participants reflect on AI's potential in teaching and learning by exploring different activities that can integrate AI literacy in education, including its ethical considerations and potential for innovative pedagogy. The approach emphasises not only acculturating professionals to AI but also empowering them to collaboratively design AI-enhanced educational activities that promote learner engagement and personalised learning experiences. Through this process, participants in the workshops develop the skills and mindset necessary to effectively leverage AI while maintaining a critical awareness of its implications in education.

## Introduction

The advent of Artificial Intelligence (AI) applications has stirred widespread interest, particularly in generative AI technologies (Chiu, 2024; Nguyen et al., 2024), not only from a professional perspective, viewing it as the Fourth Education Revolution (Seldon et al., 2020) but also raising different types of expectations from educators and learners related to the potential of AI to transform educational practices through adaptability (Gligorea et al., 2023), creative pedagogies (Romero et al., 2023), and unethical ways (Van Wyk, 2024).

Education should address the dual nature of AI technologies. While AI offers unprecedented opportunities for transformative and civic engagement through enhanced collaboration, critical and computational thinking, creativity, and problem solving (Järvela et al., 2023; Romero et al., 2023; Urmeneta & Romero, 2024), it also presents significant ethical and equity challenges. These challenges include the potential for citizen passivity and alienation as AI applications mystify and automate activities that traditionally benefit from human judgement and ethical considerations. Such a paradox underscores the importance of acculturating educators and learners to AI, ensuring they possess both the literacy and ethical frameworks needed to leverage AI positively while guarding against its less beneficial impacts. This balanced approach is crucial to realising AI's potential in education without compromising the human elements that are essential to learning and development.

**Figure 1**
Powerful technologies such as AI can be used for both good and bad.

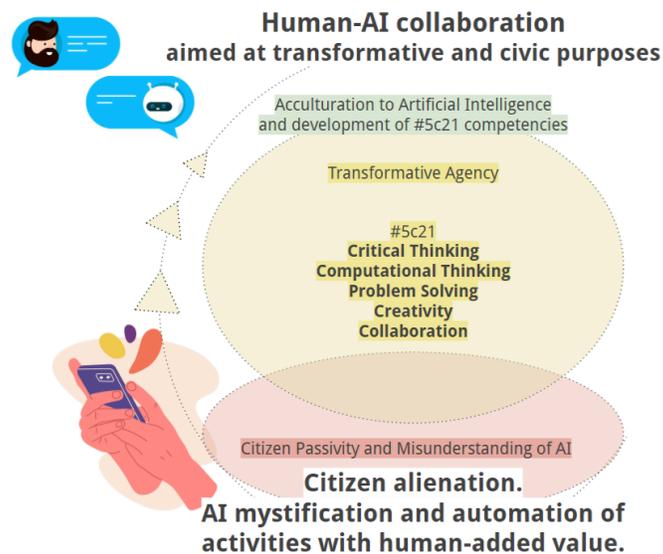

The study of AI's educational potential necessitates a comprehensive consideration within the context of technology-enhanced learning (TEL) over recent decades. Historically, technology, from headphones to personal computers and MOOCs, has carried promises of transformative educational advancements (Escueta et al., 2017; Higgins et al., 2012; Zhai et al., 2021). However, a socio-critical lens must scrutinise the optimism surrounding TEL, challenging techno-solutionist narratives that oversimplify complex societal issues (Selwin, 2023). Technological advancements, while improving processes, have prompted a nuanced understanding of technology's role in education, advocating for a balanced approach to leveraging technology for educational advancement (Elfert, 2023; Collin & Brotcorne, 2019). The adoption of AI represents a unique shift in educational stakeholder perspectives, positioning AI as a catalyst for transformative change within education (Seldon et al., 2020), aligned with the Fourth Industrial Revolution (4IR) (Zhai et al., 2021).

## Learning design support for creative pedagogy supported by AI

In this study, we develop a critical reflection on the activities developed at Terra Numerica and Maison de l'Intelligence Artificielle (MIA) and the working group Scol_IA (Romero et al., 2023; Urmeneta & Romero, 2024). The formative intervention aims to support educators in their acculturation of AI and the creative integration of AI in their educational practices. We explore how AI supports educators in developing creative pedagogical approaches through a half-day workshop hosted at MIA. The workshop is designed to provide hands-on learning experiences and collaborative design opportunities for in-service teachers and students of the MSc Smart EdTech programme at Université Côte d'Azur.

During the initial exploration phase of the workshop, participants engage in interactive activities that showcase various applications of generative and non-generative AI technologies relevant to education. These activities include demonstrations of AI-driven tools for content creation, adaptive learning systems, and intelligent tutoring systems (ITS). These activities include image generation, face and object correction, and social robotics based on AI. Through these experiences, educators gain insights into AI's potential and develop their AI literacy.

**Figure 2**
Maison de l'Intelligence Artificielle #IA06

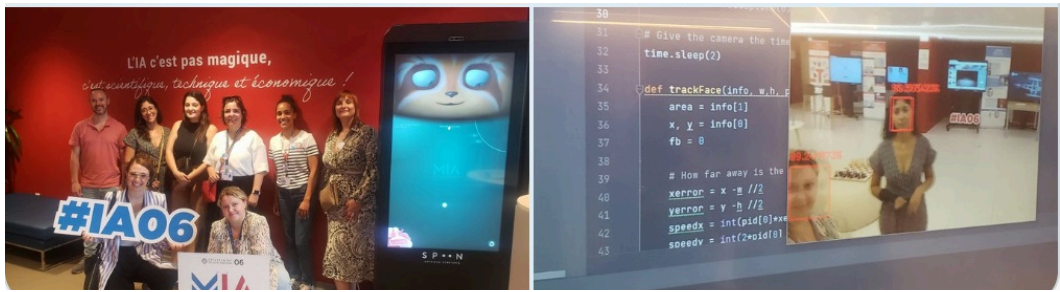

Following the exploration phase, participants transition into a team-based design exercise where they apply their understanding of AI to develop innovative pedagogical activities using the #PPAI6 model framework. This model encourages educators to leverage AI for creative purposes in education, such as fostering critical thinking, promoting personalised learning pathways, or facilitating collaborative projects. By providing practical guidance and collaborative opportunities, this workshop bridges the gap between AI technology and effective educational practice. The results of our research provide valuable insights and resources to support the ongoing integration of AI in education, facilitating continuous improvement and adaptation in pedagogical approaches.

## Human-AI collaboration in education through a creative pedagogy perspective

With the increasing prevalence and accessibility of AI systems, the dynamics between humans and AI have progressed from mere process automation to a collaborative partnership founded on mutual synergies and strengths (Razmerita et al., 2022), as well as the metacognitive potential inherent in human-AI collaboration (Romero et al., 2023). It is within this intersection of human intuition and imagination, coupled with AI's computational power and processing capabilities, that the greatest potential for human-AI co-creativity emerges. This potential is already evident across various creative domains, such as music composition and performance (Rohrmeier, 2022), as well as in the visual arts, where artists have utilized AI tools to enrich their creative processes (Kim et al., 2021), leveraging an extensive learning approach aimed at transformative objectives (Romero et al., 2023). However, for our focus, it is the prospect of reimagining creativity through the prism of human-AI interactions, innovative teaching strategies, and learner-centric activities that may enable us to nurture learners' agency and foster a creative pedagogy supported by human-AI collaborations. To facilitate the identification of the different levels of creative engagement in the use of AI, the Passive-Participatory (PP) model for AI in education (#PPai6) distinguishes six levels of creative engagement (Figure 3).

**Figure 3**
Creative engagement in AI in education

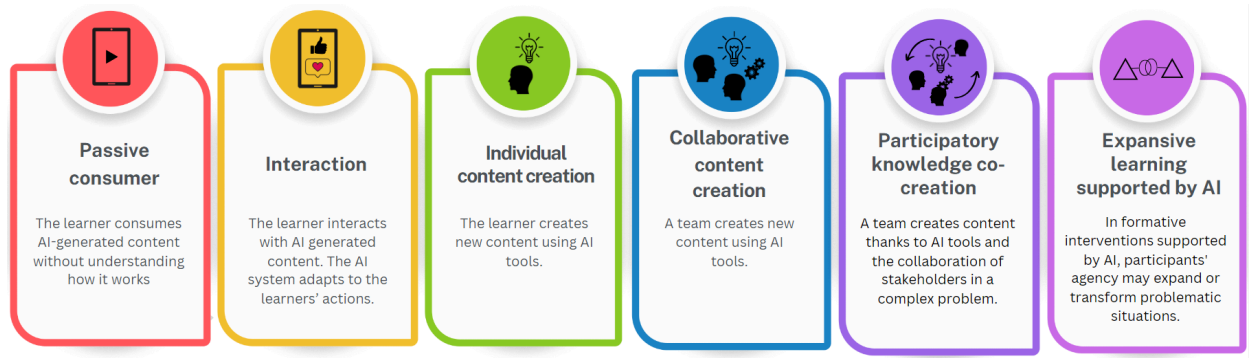

At the first level, learners act as passive consumers, engaging with AI-generated content without a full understanding of its workings. Moving through the levels, learners progress to become interactive consumers, actively interacting with AI-generated content as the AI system adapts to their actions. Levels three and four involve individual and collaborative content creation, respectively, with learners utilising AI tools to generate new content. The fifth level, participatory knowledge co-creation, sees teams creating content with the aid of AI tools and collaboration from stakeholders to tackle complex problems. At the sixth and most advanced level, expansive learning supported by AI, participants' agency expands or transforms problematic situations through formative interventions. AI tools play a crucial role in identifying contradictions in complex problems, generating concepts or artefacts to regulate conflicting stimuli, and fostering collective agency and action. While the potential for AI to reach this transformative level is immense, it is noteworthy that the majority of current AI in education studies operate at the second level (interactive consumer), primarily relying on Intelligent Tutoring Systems (ITS). The exploration of higher levels presents an exciting frontier for the future development and implementation of AI in education that will be explored through the different case studies of this book.

To facilitate the design, we propose that pre-service teachers and edtech professionals develop their co-design learning activity using a template. The template provides a structured approach for designing learning activities that effectively integrate AI into education as a support for the learners activities or as a way to develop their learners' AI literacy (Alexandre et al., 2020). The formative intervention follows a step-by-step process for defining the learning objectives, targeting specific learner groups, and outlining the entire process from conception to evaluation.

- Step 1 invites users to choose and articulate a learning objective for their activity, probing the rationale behind the selection to ensure alignment with broader educational goals.
- Step 2, termed "Learning Activity," details the breakdown of the activity into three stages: Preparation, Development, and Debriefing. This sequential structuring is crucial for organizing content and interaction phases efficiently.
- Step 3 addresses the resources necessary for the activity, emphasizing the importance of both analogical and technological tools to foster a conducive learning environment.
- Step 4 encourages users to describe specific learning strategies that will be employed during the activity, tailoring approaches to enhance engagement and effectiveness.
- Step 5 allows for the naming of the activity and the specification of learning modalities (online, face-to-face, or hybrid), crucial for setting the context in which the activity will be executed. Additionally, it prompts the designer to consider potential success factors and challenges during the activity orchestration.

**Figure 4**
Template for the co-design of an AI-enhanced learning activity.

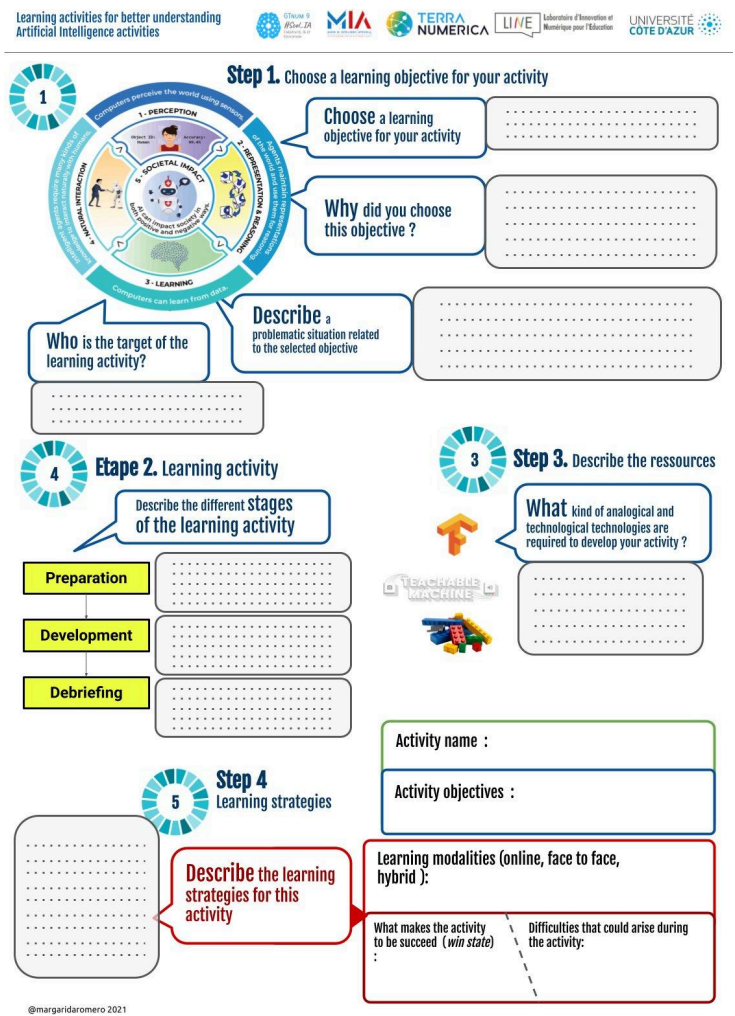

The use of the template has permitted a reduction of the time required to develop a first draft of an AI-enhanced learning activities for their educational context, but also a better focus during the coordination efforts of the co-design activity.

**Figure 5**

Students engaged in the co-design of an AI-enhanced learning activity.

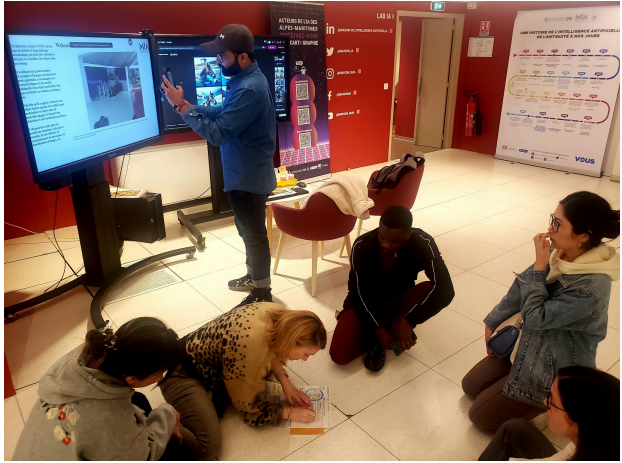

## Discussion

Education is undergoing a profound transformation as we embrace the potential of AI to enhance creative pedagogies, as evidenced by our recent workshop at Maison de l'Intelligence Artificielle (MIA). Our workshop highlighted that education extends beyond the transmission of knowledge; it aims to cultivate creative thinking, critical skills, and a lifelong love of learning. Human-AI collaborations that support and augment learners and educators activities require the development of AI literacy and supporting the complexity of learning design activities (Duret & Romero, 2022) through visualisation tools such as the co-design template proposed for this formative intervention. This transformative approach presents an opportunity to redefine creative pedagogies by integrating AI into educational practices across different levels and domains, thus opening up new avenues for personalised, engaging, and transformative learning experiences. By adopting an expansive learning approach (Engeström & Sannino, 2010; Romero, Duguay, et al., 2023), we can develop educators' agencies to identify and integrate AI's capabilities to augment teaching and learning activities while supporting the development of AI literacy for everyone.

## References

Alexandre, F., Becker, J., Comte, M. H., Lagarrigue, A., Liblau, R., Romero, M., & Viéville, T. (2020). *Open Educational Resources and MOOC for Citizen Understanding of Artificial Intelligence*. Inria.
Chiu, T. K. (2024). Future research recommendations for transforming higher education with generative AI. *Computers and Education: Artificial Intelligence*, 6, 100197.
Collin, S., & Brotcorne, P. (2019). Capturing digital (in) equity in teaching and learning: A sociocritical approach. *The International Journal of Information and Learning Technology*, 36(2), 169-180.

Duret, C., & Romero, M. (2022). L'activité de conception comme démarche créative pour la formation des enseignants à l'intégration du numérique dans l'activité d'enseignement-apprentissage. *Revue internationale du CRIRES: innover dans la tradition de Vygotsky*, 6(3), 46-65.

Elfert, M. (2023). Humanism and democracy in comparative education. *Comparative Education*, 1-18.

Escueta, M., Quan, V., Nickow, A. J., & Oreopoulos, P. (2017). *Education Technology: An Evidence-Based Review* (Working Paper 23744). National Bureau of Economic Research. https://doi.org/10.3386/w23744

Higgins, S., Xiao, Z., & Katsipataki, M. (2012). The Impact of Digital Technology on Learning: A Summary for the Education Endowment Foundation. Full Report. In *Education Endowment Foundation*. Education Endowment Foundation. https://eric.ed.gov/?id=ED612174

Gligorea, I., Cioca, M., Oancea, R., Gorski, A. T., Gorski, H., & Tudorache, P. (2023). Adaptive Learning Using Artificial Intelligence in e-Learning: A Literature Review. *Education Sciences*, 13(12), 1216.

Järvelä, S., Nguyen, A., & Hadwin, A. (2023). Human and artificial intelligence collaboration for socially shared regulation in learning. *British Journal of Educational Technology*, 54(5), 1057-1076.

Nguyen, A., Hong, Y., Dang, B., & Huang, X. (2024). Human-AI collaboration patterns in AI-assisted academic writing. *Studies in Higher Education*, 1-18.

Romero, M., Isaac, G., Barma, S., Girard, M. A., & Heiser, L. (2023). Critical thinking, creativity, and agency for the development of regenerative cultures. IRMBAM. https://hal.science/hal-04593519

Romero, M., Heiser, L., & Lepage, A. (2023). *Enseigner et apprendre à l'ère de l'intelligence artificielle: acculturation, intégration et usages créatifs de l'IA en éducation: livre blanc*. Canopé.

Seldon, A., Abidoye, O., & Metcalf, T. (2020). *The Fourth Education Revolution Reconsidered: Will Artificial Intelligence Enrich Or Diminish Humanity?*. Legend Press Ltd.

Selwyn, N. (2023). Lessons to be learnt? Education, techno-solutionism, and sustainable development. *Technology and Sustainable Development*, 71.

Urmeneta, A., & Romero, M., (2024). *Creative Applications of Artificial Intelligence in Education*. Palgrave.

Van Wyk, M. M. (2024). Is ChatGPT an opportunity or a threat? Preventive strategies employed by academics related to a GenAI-based LLM at a faculty of education. *Journal of Applied Learning and Teaching*, 7(1).

Zhai, X., Chu, X., Chai, C. S., Jong, M. S. Y., Istenic, A., Spector, M., Liu, J.-B., Yuan, J., & Li, Y. (2021). A Review of Artificial Intelligence (AI) in Education from 2010 to 2020. *Complexity*, *2021*, e8812542. https://doi.org/10.1155/2021/8812542